\documentclass[runningheads]{llncs}
\usepackage{ulem}
\usepackage{graphicx}
\usepackage{amsmath}
\usepackage{amssymb}
\usepackage{amsfonts}
\usepackage{xcolor}
\usepackage{enumerate}
\usepackage{algorithm}
\usepackage{algorithmic}

\usepackage{subcaption} 

\begin{document}
\title{LaMoD: Latent Motion Diffusion Model For Myocardial Strain Generation}


%
%
\author{Jiarui Xing\inst{1*} \and
Nivetha Jayakumar\inst{1}\and 
Nian Wu\inst{1} \and
Yu Wang\inst{3} \and
Frederick H. Epstein\inst{3} \and
Miaomiao Zhang\inst{1,2}}

\authorrunning{F. Author et al.}

\institute{Department of Electrical and Computer Engineering, University of Virginia, USA \and
Department of Computer Science, University of Virginia, USA \and
Department of Biomedical Engineering, University of Virginia Health System, USA 
\email{jx8fh@virginia.edu}\\
}
\maketitle              
\begin{abstract}
Motion and deformation analysis of cardiac magnetic resonance (CMR) imaging videos is crucial for assessing myocardial strain of patients with abnormal heart functions. Recent advances in deep learning-based image registration algorithms have shown promising results in predicting motion fields from routinely acquired CMR sequences. However, their accuracy often diminishes in regions with subtle appearance changes, with errors propagating over time. Advanced imaging techniques, such as displacement encoding with stimulated echoes (DENSE) CMR, offer highly accurate and reproducible motion data but require additional image acquisition, which poses challenges in busy clinical flows. In this paper, we introduce a novel Latent Motion Diffusion model (LaMoD) to predict highly accurate DENSE motions from standard CMR videos. More specifically, our method first employs an encoder from a pre-trained registration network that learns latent motion features (also considered as deformation-based shape features) from image sequences. Supervised by the ground-truth motion provided by DENSE, LaMoD then leverages a probabilistic latent diffusion model to reconstruct accurate motion from these extracted features. Experimental results demonstrate that our proposed method, LaMoD, significantly improves the accuracy of motion analysis in standard CMR images; hence improving myocardial strain analysis in clinical settings for cardiac patients. Our code is publicly available at \texttt{https://github.com/jr-xing/LaMoD}.
 
\end{abstract}
\section{Introduction}
Motion and deformation analysis of CMR videos provides valuable measurements to quantify myocardial strain in patients with heart disease, offering clinically significant data for disease assessment, diagnosis, and treatment planning~\cite{popovic2008association,balter2007imaging,seo2013computing,xing2021deep,xing2023multitask}. CMR feature tracking (FT) is widely used in clinical practice to evaluate myocardial motion, deformation, and strain functions~\cite{tee2013imaging,morales2021deepstrain,qiao2020temporally,qin2018joint}. Such a technique employs optical flow-based algorithms~\cite{horn1981determining} to track image features or patterns within the myocardium throughout the cardiac cycle. While FT is convenient and integrates well into routine clinical workflows, this technique is generally less accurate due to its limited motion tracking accuracy~\cite{amzulescu2019myocardial,young2012generalized}.

With recent advancements in deep learning, several research groups have utilized image registration-based networks to predict myocardial motion and strain from CMR images~\cite{morales2021deepstrain,qiao2020temporally,xing2024multimodal}. Existing methods incorporated new regularization techniques, such as enforcing temporal consistency~\cite{qiao2020temporally}, or introducing prior knowledge of cardiac biomechanics ~\cite{qin2020biomechanics,zhang2022learning,qin2023generative}, to achieve improved motion prediction quality. However, these approaches still partially address the challenges of detecting motion in regions with subtle image appearance changes (i.e., the myocardium mid-wall motion)~\cite{amzulescu2019myocardial,claus2015tissue,pedrizzetti2016principles}, leading to compromised accuracy of myocardial strain measurements. To alleviate this issue, another research line has utilized ``ground-truth'' data from the advanced imaging technique DENSE, which provides highly accurate and reproducible myocardial motion data to supervise the motion learning process~\cite{wang2023strainnet,wang2024transstrainnet}. In contrast to conventional CMR techniques, DENSE directly encodes tissue displacement within the imaging data, allowing for precise quantification of myocardial motion throughout the cardiac cycle. Despite promising progress in predicting myocardial motion for improved strain analysis under the guidance of DENSE~\cite{wang2023strainnet,wang2024transstrainnet}, current approaches heavily rely on features extracted from segmented myocardial contours rather than the underlying cardiac motions, potentially compromising strain accuracy. Furthermore, the network training involved only DENSE contours that may not fully generalize to standard CMR images.

To address this problem, we propose to develop a novel Latent Motion Diffusion model (LaMoD) to further improve the current deep networks to predict highly accurate DENSE motions from standard CMR videos. In particular, our method, LaMoD, first employs an encoder from a pre-trained registration network that learns latent motion features of myocardial deformations derived from both cine and DENSE image sequences. Supervised by the ground-truth motion provided by DENSE, LaMoD then leverages a probabilistic latent diffusion model to reconstruct accurate motion from
these extracted features. Once our model is trained, the DENSE data is no longer required in the testing phase. Our contributions are threefold:
\begin{enumerate}[(i)]
\item Develop a new framework, LaMoD, that generates time-sequential myocardial deformation fields from a learned latent space of motion features.
\item Our method is the first to leverage latent diffusion models in the motion space to produce highly accurate myocardial strain from standard CMR videos. 
\item Opens promising research avenues for transferring knowledge from advanced strain imaging to routinely acquired CMR data; hence maximizing benefits for patients with cardiac diseases.
\end{enumerate}

\section{Background: Registration-based Motion Deformation Estimation}
In this section, we briefly review the concept of image registration, which is a fundamental technique for estimating motion deformation between images. We will apply this concept to motion tracking in CMR video sequences.

Given a source image $S$ and a target image $I$, the problem of diffeomorphic image registration is typically formulated as an energy minimization over a time-dependent deformation fields, $\{\phi_t:t\in[0,1]\}$, i.e.,
\begin{equation}
E(\phi_1) = \frac{1}{2\sigma^{2}}\, \text{Dist}(S \circ \phi_1^{-1}, I) + \text{Reg}(\phi_1).
\label{eq:TotalEnergy}
\end{equation}
Here, $\circ$ denotes an interpolation operator, which deforms a source image $S$ to match the target $I$. The Dist(·,·) is a distance function that measures the dissimilarity between images weighted by a positive parameter $\sigma$, and Reg($\cdot$) is a regularization term to enforce the smoothness of transformation fields. In this paper, we use a commonly used sum-of-squared intensity differences ($L_2$-norm)~\cite{beg2005computing} as the distance function.

In our implementation, we adopt the large diffeomorphic deformation metric mapping (LDDMM) framework~\cite{beg2005computing,zhang2015fast} to generate diffeomorphic deformations, parameterized by time-dependent velocity fields, $v_t: t \in [0, 1]$, i.e., 
\begin{eqnarray}
\label{eq:velocity}
\frac{d\phi_t}{dt} = v_t \circ \phi_t.
\end{eqnarray}

A geodesic shooting algorithm~\cite{miller2004computational,vialard2012diffeomorphic} has shown that the geodesic path of deformation, $\phi_t$, with a given initial condition, $v_0$, can be uniquely determined through integrating the Euler-Poincaré differential equation (EPDiff)~\cite{arnold1966,miller2006} as 
\begin{align}\label{eq:epdiff}
\frac{d v_t}{dt} = -K \left[\left( D v_t\right)^\mathrm{T} m_t + D m_t \, v_t + m_t \operatorname{div} v_t \right],
\end{align}
where $D$ denotes a Jacobian matrix and $\rm{div}$ is the divergence operator. The $K$ denotes an inverse operator of $L: V \rightarrow V^*$, which is a symmetric, positive-definite differential operator that maps a tangent vector $v \in V$ into the dual space $m \in V^*$. Numerous studies and applications have used the initial velocities, $v_0$, to represent diffeomorphic deformations in the context of motion/deformation analysis~\cite{zhang2014bayesian,miller2004computational}.

\section{Our Model: LaMoD}
\begin{figure*}[!ht]
\centering
\includegraphics[width=1.0\textwidth] {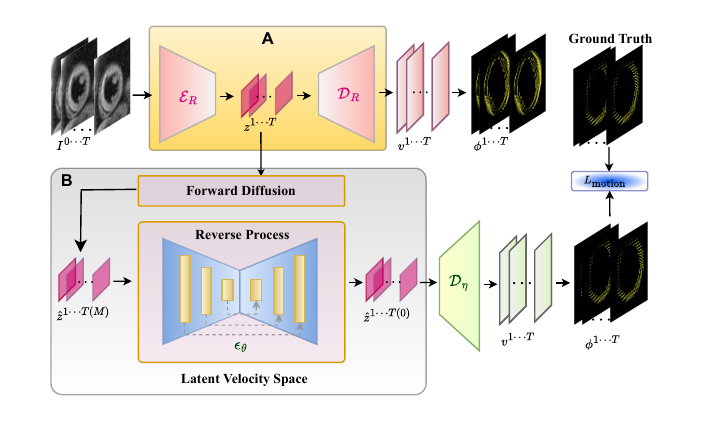}
     \caption{An overview of our proposed network framework. (A) Registration-based network to learn latent motion features represented by initial velocity fields. (B) Diffusion model in latent motion spaces.}
\label{fig:NetArch}
\end{figure*}

Our model consists of two main components: (i) a pre-trained registration network based on the LDDMM framework~\cite{beg2005computing} to extract latent motion features from CMR sequences, and (ii) a motion prediction model leveraging latent diffusion models to reconstruct realistic and highly accurate myocardial motion, supervised by DENSE data. An overview of our proposed framework, LaMoD, is illustrated in Fig.~\ref{fig:NetArch}.

\subsection{Latent Motion Feature Learning From Image Videos}

Given a image sequence, $\{I^{\tau}\}^T_{\tau=0}$, that includes $T+1$ time frames covering an entire motion cycle of myocardium.
We pre-train a LDDMM-based registration network~\cite{Hinkle2021jacobhinkle} to learn the deformation fields between the initial frame $I^0$ and each subsequent frame $\{I^{\tau}\}$. This results in a number of $T$ pairs of images, i.e., $\{(I^0, I^1), (I^0, I^2) \cdots, (I^0, I^T)\}$. To utilize the intrinsic spatial connections and motion consistency across the sequence of time frames, we stack the $T$ pairwise images into a 3D volume and predict the deformation fields simultaneously in the implementation. We employ an UNet architecture~\cite{ronneberger2015unet} as our network backbone, featuring an encoder $\mathcal{E}_R$ that maps the input image pairs $\{(I^0, I^{\tau})\}$ to the corresponding latent velocity features $\{z^{\tau}\}$, and a decoder $\mathcal{D}_R$ that project $\{z^{\tau}\}$ back to the input image space, i.e., initial velocity fields $\{v_0^{\tau}\}$. The corresponding final transformation fields, $\{\phi_1^{\tau}\}$, are computed through Eq.~\eqref{eq:velocity} and Eq.~\eqref{eq:epdiff}. For a simplified notation, we will drop the time index in following sections, i.e., $\phi^{\tau}_1 \overset{\Delta}{=} \phi^{\tau}$ and $v^{\tau}_0 \overset{\Delta}{=} v^{\tau}$.

\subsection{Diffusion Model In Latent Motion Spaces}
We introduce a new latent motion diffusion module that captures the complex distribution of latent motion features, resulting in higher quality motion reconstruction under the supervision of DENSE ground-truth. More specifically, the registration-based latent velocity features are first refined through a diffusion process and then fed into a motion reconstruction network to predict the final, highly accurate myocardial motion. This approach ensures a more precise and realistic depiction of myocardial dynamics, demonstrating significant improvements over existing methods.

Inspired by the Denoising Diffusion Probabilistic Models (DDPM)\cite{ho2020denoising,rombach2021highresolution}, we formulate the model as a latent Markov chain with $M$ steps. The framework has a forward and reverse diffusion process. The forward process iteratively adds random Gaussian noise to the input latent features over $m \in [1,2,...,M]$ steps. Similar to~\cite{jayakumar2023sadir}, we employ smoothed Gaussian noise $\epsilon'$ in the diffusion process to facilitate faster optimization convergence compared to normal Gaussian noise $\epsilon$, i.e. $\epsilon'=\mathcal{K}(\epsilon)$, where $\epsilon\sim\mathcal{N}(0,\mathbf{I})$ and $\mathcal{K}(\cdot)$ is a Gaussian smoothing kernel.

Defining $\boldsymbol{z}^{(m)} \triangleq \{z^{1(m)},z^{2(m)},\dots,z^{T(m)}\}$ as the learned latent motion features (represented by initial velocities) for all frames at step $m$, our {\bf forward diffusion process} is defined as:
\begin{equation}
    q(\boldsymbol{z}^{(m)} | \boldsymbol{z}^{(m-1)}) = \mathcal{N}(\boldsymbol{z}^{(m)}; \sqrt{1-\beta_m}\boldsymbol{z}^{(m-1)}, \beta_m \mathbf{I})
    \label{eq:dif-forward}
\end{equation}
where $\beta_m$ is a time-dependent variance schedule used to parameterize the probabilistic transitions. The forward process starts from the registration-based latent feature, i.e. $\boldsymbol{z}^{(0)}=\{z^\tau\}=\{z^1,z^2,\dots,z^T\}$. Alternatively, the forward diffusion process can be formulated as a single step process:
\begin{gather}
    q(\boldsymbol{z}^{(m)} | \boldsymbol{z}^{(0)}) = \mathcal{N}(\boldsymbol{z}^{(m)}; \sqrt{\bar{\alpha}_m}\boldsymbol{z}^{(0)}, (1-\bar{\alpha}_m)\mathbf{I}) \label{eq:dif-forward-onestep-prob}\\
    \boldsymbol{z}^{(m)} = \sqrt{\bar{\alpha}_m}\boldsymbol{z}^{(0)} + \sqrt{1-\bar{\alpha}_m} ~\mathcal{K}(\epsilon), \label{eq:dif-forward-onestep-det}
\end{gather}
where $\bar{\alpha}_m = \prod_{i=1}^m \alpha_i$ and $\alpha_m = 1- \beta_m$. 

The reverse diffusion process aims to restore the latent feature with a deep neural network. It takes place over $m$ steps as:
\begin{equation}
    p_\theta(\boldsymbol{\hat{z}}^{(m-1)}|\boldsymbol{\hat{z}}^{(m)}) = \mathcal{N}(\boldsymbol{\hat{z}}^{(m-1)};\mu_\theta(\boldsymbol{\hat{z}}^{(m)},m), \Sigma_\theta(\boldsymbol{\hat{z}}^{(m)},m)), 
\end{equation}
where $\boldsymbol{\hat{z}}^{(M)}$ are the latent velocity features obtained from the reverse process, $\mu_\theta = \frac{1}{\sqrt{\alpha_m}}(\boldsymbol{\hat{z}}^{(m)} - \frac{\beta_m}{\sqrt{1-\bar{\alpha}_m}}\epsilon_\theta(\boldsymbol{\hat{z}}^{(m)},m))$ and $\Sigma_\theta(\boldsymbol{\hat{z}}^{(m)},m)= \beta_m\mathbf{I}$ are the mean and variance of the transitions of the reverse process respectively.


The {\bf reverse process} is implemented by training a noise-prediction network $\epsilon_{\theta}$ that takes the noisy motion feature $\boldsymbol{\hat{z}}^{(m)}$ and the step $m$ as input. For each step $m$, the sampling in the reverse process is defined as:
\begin{gather}
    \boldsymbol{\hat{z}}^{(m-1)} = \frac{1}{\sqrt{\alpha_m}} (\boldsymbol{\hat{z}}^{(m)} - \frac{1-\alpha_m}{\sqrt{1-\bar{\alpha}_m}}\epsilon_\theta(\boldsymbol{\hat{z}}^{(m)},m)) + \sigma_m \mathcal{K}(\gamma) ,
    \label{eq:dif-back}
\end{gather}
where $\gamma \sim \mathcal{N}(0, \mathbf{I})$.

Noting the shared diffusion step as $m$, and latent features as well as the additive noise of the $n$-th input sequence as $\boldsymbol{z}_n^{(m)}$ and $\epsilon'_n$, respectively, the corresponding loss function over the whole training dataset with $N$ sequences is defined as
\begin{equation}
    l_{\text{diffusion}} = \sum_{n=1}^N{E_{m,z,\epsilon}\left[\|\epsilon'_n-\epsilon_\theta(\boldsymbol{z}_n^{(m)},m))\|_2\right]} + \lambda_\epsilon\text{reg}(\theta),
    \label{eq:diffusionloss}
\end{equation}
where the function $\text{reg}(\cdot)$ represents network $L_2$ weight decay regularity weighted by $\lambda_{\epsilon}$. 

The refined latent features $\boldsymbol{z}^{(0)}$ are fed into to the motion decoder, $\mathcal{D}_\eta$, to reconstruct the accurate myocardial motion of original data resolution, i.e., $\hat{\boldsymbol{\Phi}}_n=\{\hat{\Phi}^1_n,\hat{\Phi}^2_n,\dots,\hat{\Phi}^T_n\}=\mathcal{D}_\eta(\boldsymbol{z}^{(0)}_n)$, where $\eta$ is the motion decoder and $n \in \{1, \cdots, N\}$ is the data index. With the supervision of DENSE ground truth, $\boldsymbol{\Phi}_n$, our motion reconstruction loss is defined as
\begin{align}
    l_{\text{motion}} 
    &=\frac{1}{N}\sum_{n=1}^N\sum_{\tau=1}^T \| \Phi_{n}^{\tau}-\hat{\Phi}_{n}^{\tau} \|^2_2 + \lambda_M\text{reg}(\eta), \nonumber \\
    &= \frac{1}{N}\sum_{n=1}^N\sum_{\tau=1}^T \| \Phi_{n}^{\tau}-\mathcal{D}_\eta(\boldsymbol{z}^{(0)}_n) \|^2_2 + \lambda_M\text{reg}(\eta), 
    \label{eq:motionloss}
\end{align}
where $\lambda_M$ is the weighting parameter of weight decay regularization term $\text{reg}(\cdot)$. 


We jointly train the loss functions of diffusion model (Eq.~\eqref{eq:diffusionloss}) and the motion reconstruction (Eq.~\eqref{eq:motionloss}) till its convergence. We define the total loss as
\begin{equation*}
      l_{\text{total}} = l_{\text{diffusion}} + \alpha l_{\text{motion}},
\end{equation*}
where $\alpha$ is the loss weighting term. We summarize our model in Algorithm 1.
\begin{algorithm}[!t]
\caption{LaMoD: Latent Motion Diffusion Model For Myocardial Strain Generation}
\begin{algorithmic}[1]
\REQUIRE DENSE CMR videos, $\{I^\tau\}$, with directly encoded motion, $\{\Phi^\tau\}$;
\ENSURE Reconstructed myocardial motion, $\{\hat{\Phi}^\tau\}$;
\vspace{0.5em}

\textbf{Latent Motion Feature Learning}
\STATE $\boldsymbol{z}^{(0)} = \{z^1,\dots,z^T\} = \mathcal{E}_R(\{I^0,\dots,I^T\})$
\vspace{0.5em}

\textbf{Diffusion Model Inference}
\STATE $\epsilon\sim\mathcal{N}(0,\mathbf{I})$
\STATE $\boldsymbol{z}^{(M)} = \sqrt{\bar{\alpha_{M}}}\boldsymbol{z}^{(0)} + \sqrt{1-\bar{\alpha_{M}}} ~\mathcal{K}(\epsilon)$
\FOR {$m=M, M-1,\dots,1$}
    \STATE $\boldsymbol{\hat{z}}^{(m-1)} = \frac{1}{\sqrt{\alpha_m}} (\boldsymbol{\hat{z}}^{(m)} - \frac{1-\alpha_m}{\sqrt{1-\bar{\alpha_m}}}\epsilon_\theta(\boldsymbol{\hat{z}}^{(m)},m)) + \sigma_m \mathcal{K}(\gamma)$
\ENDFOR
\vspace{0.5em}

\STATE $\{\hat{\Phi}^\tau\} = \mathcal{D}_\eta(\boldsymbol{\hat{z}}^{(0)})$; 

\RETURN $\{\hat{\Phi}^\tau\}$
\label{alg}
\end{algorithmic}
\end{algorithm}


\section{Experimental Evaluation}
We first demonstrate the effectiveness of LaMoD on DENSE CMR videos and then test its performance on standard cine CMRs, highlighting its clinical potential. Both quantitative and visualization results are presented. Note that our network is trained solely on the DENSE dataset and then tested on both DENSE and CINE datasets.

The training of all our experiments is implemented on an server with AMD EPYC $7502$ CPU of $126$GB memory and Nvidia GTX $3090$Ti GPUs. We train our networks using Adam optimizer~\cite{kingma2014adam} with maximal $2000$ epochs with the early stop strategy. The batch size is set to $32$ and weight decay weights are set to $\lambda_\epsilon=\lambda_M=1E-4$. The hyper-parameters are optimized with grid search strategy. The optimal learning rate is $1E-4$ and the optimal loss weight is $\alpha=1E-2$.



\subsection{Dataset}
\noindent {\bf DENSE CMR videos with directly encoded motion data.} We utilize $741$ DENSE CMR videos of the left ventricular (LV) myocardium collected from $284$ subject, including 124 healthy volunteers and 160 patients with various types of heart disease. Data are collected from eight centers (University of Virginia, Charlottesville; University Hospital, Saint-Etienne, France; University of Kentucky, Lexington; University of Glasgow, Scotland; St Francis Hospital, New York; the Royal Brompton Hospital, London, England; Emory University, Atlanta, Georgia; and Stanford University, Palo Alto, California). Each DENSE scan was performed in $4$ short-axis planes at the basal, two midventricular, and apical levels (with  temporal resolution of $17$ ms, pixel size of $2.65^2 \text{ mm}^2$, and slice thickness=$8$mm). Other parameters include displacement encoding frequency $k_e = 0.1 \text{ cycles/mm}$, flip angle $15^{\circ}$, and echo time $= 1.08 \text{ ms}$.  \\

\noindent {\bf Standard cine CMR videos.} We tested our proposed model on $105$ short-axis cine CMRs slices of $40$ subjects from the DENSE dataset mentioned above, including $14$ patients and $26$ volunteers. 
All scans were acquired during repeated breath hold covering the left ventricle (LV) (field of view, 320 × 320 to 380 × 380 mm2; temporal resolution, 30–55 msec, depending on heart rate). Each selected cine slice corresponds to a DENSE scan of the same patient at the same spatial location (within $\pm2\text{mm}$), allowing the DENSE displacement field to serve as the ground-truth motion for the cine slices. \\

\noindent {\bf Data Pre-processing.} All cine and DENSE CMR sequences were temporally and spatially aligned for efficient network training. In particular, the standard cine sequences were temporally resampled to $40$ frames to match the DENSE temporal resolution. All images were resampled to a \(1 \text{ mm}^2\) resolution and cropped to \(128 \times 128\). We ran all experiments on binary LV myocardium segmentation from both cine and DENSE sequences to avoid appearance gaps between the magnitude images, using masks manually labeled by clinical experts.

\subsection{Experimental Design}
We used all standard CMR and DENSE videos to train our registration network that can effectively learn latent motion features. However, we only use DENSE motion to train the diffusion model to reconstruct myocardial deformations. In particular, the DENSE data set was divided into 538 samples for training, 101 for validation and 102 for testing. We employed a site-balanced splitting strategy, maintaining consistent proportions from each site across all sets to ensure representative sampling and mitigate site-specific biases. Paired DENSE data with cine CMR videos were used for regional segmental strain comparison in our experimental evaluation.\\ 


\noindent {\bf Evaluate motion and strain error on DENSE data.} We first evaluate the quality of predicted motions for DENSE sequences, using pixel-wise motion field error (i.e., end-point error (EPE)) defined as the Euclidean distance between the predicted and ground truth motion vectors. We compare the performance of our proposed model with three state-of-the-art deep learning-based motion/strain prediction models — StrainNet~\cite{wang2023strainnet}, UNetR~\cite{hatamizadeh2022unetr}, 3D TransUNet~\cite{chen20233d}. All methods are trained on the same dataset, and their best performances are reported.   

We then generate strain maps from the predicted motion fields for the DENSE input and compare it with all baseline algorithms. We performed a quantitative analysis of strain errors on regional segmental strains. Consistent with previous studies~\cite{wang2023strainnet,wang2024transstrainnet}, we divided the myocardium in each DENSE slice into six segments, starting from the right ventricle insertion point and proceeding counterclockwise. We then calculated the average absolute error for strain evaluation in each segment. \\


\noindent {\bf Evaluate strain error on standard cine CMR videos.} 
For cine sequences, our evaluation focuses exclusively on segmental strain error due to the significant differences between the paired DENSE contours and the collected cine CMR myocardium contours. This is mainly due to the spatial misalignment between the myocardium regions from input cine images and the DENSE-derived ground truth, which are collected from separate scans, making pixel-wise error computation impractical. We apply the same segmental strain approach as used for the DENSE data, dividing the myocardium into six segments and calculating the average absolute strain error for each. Furthermore, we compared strain values with those obtained using widely used commercial software (SuiteHeart version 5.0.4; NeoSoft) based on Feature Tracking (FT) in clinical settings.



\subsection{Experimental Results}
\begin{figure*}[!ht]
    \centering
    \includegraphics[width=1.0\linewidth]{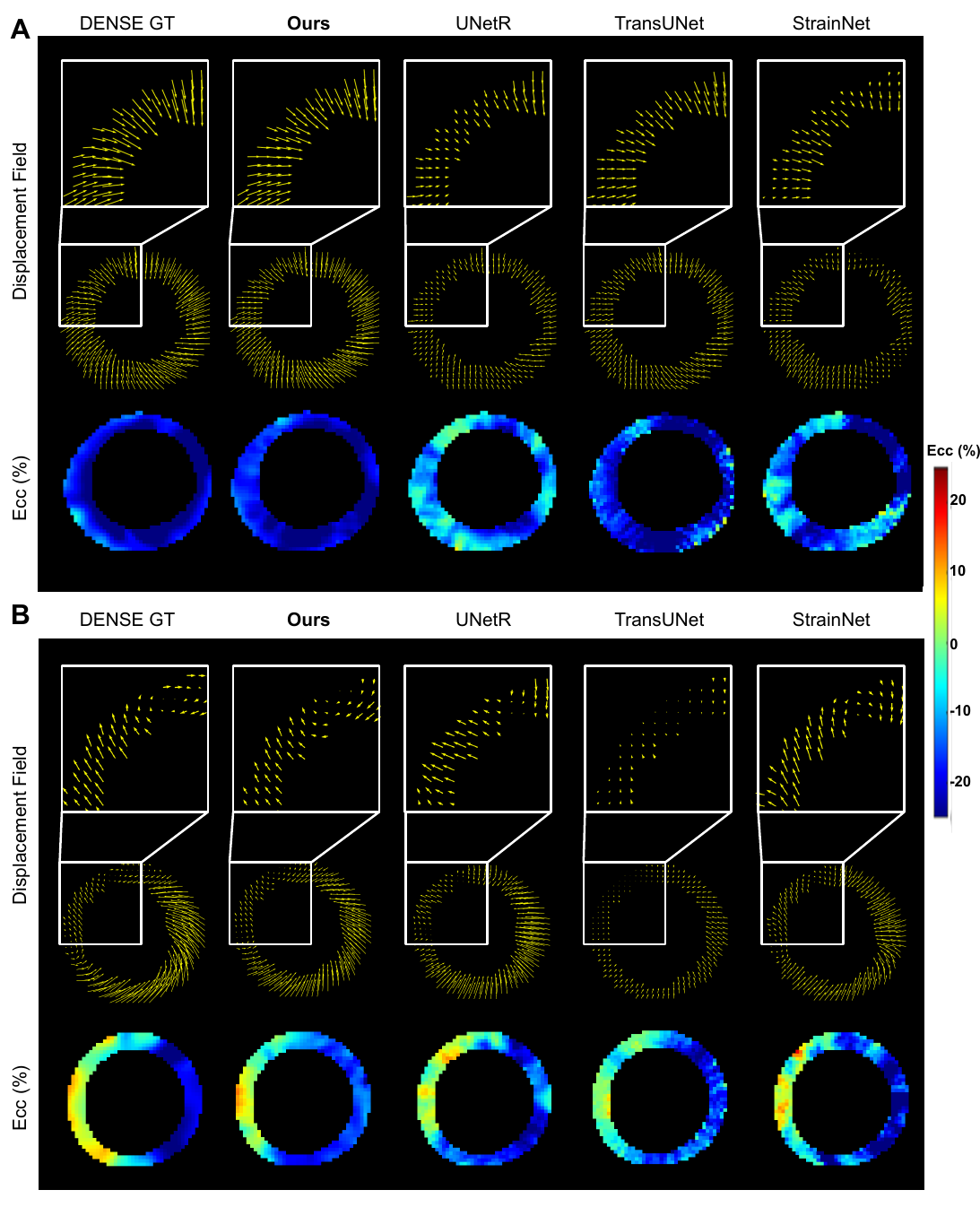}
    \caption{Panel A (healthy volunteer) \&  B (heart failure patient with left bundle branch block): exemplary comparison of end-systolic displacement and circumferential strain maps (Ecc) derived from DENSE input across all methods. Left to right: DENSE ground truth and predictions from our model vs. baselines. Top to bottom: enlarged view of selected displacement region; full displacements; circumferential strain maps (contraction in blue vs. stretch in red).} 
    \label{fig:DENSE-example}
\end{figure*}
Fig.~\ref{fig:DENSE-example} (test on DENSE data) presents the visualizations of the predicted end-systolic displacement fields and the circumferential (Ecc) strain  maps for all models compared to the ground truth from DENSE ground truth. To provide a comprehensive evaluation, we include examples from both healthy volunteers (top panel: ground A) and patients with heart failure (bottom panel: ground B), specifically those with left bundle branch block (LBBB). 
The predictions generated by our method consistently show a closer resemblance to the ground truth across all cases. These results collectively demonstrate that our method provides more accurate and robust performance compared to the baseline models.

\begin{figure*}[!ht]
    \centering    
    \includegraphics[width=1.0\linewidth]{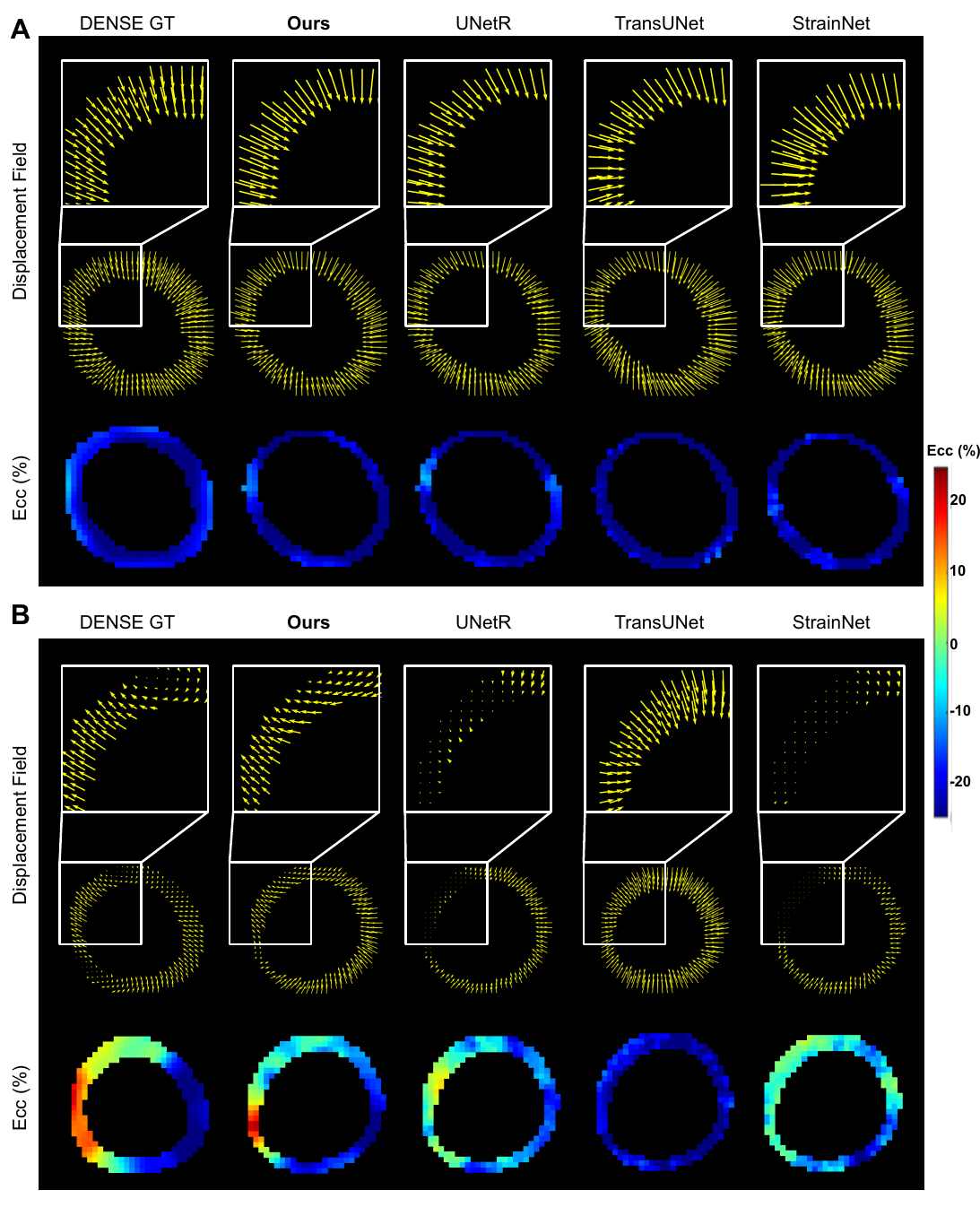}
    \caption{Panel A (healthy volunteer) \&  B (heart failure patient with left bundle branch block): exemplary comparison of end-systolic displacement and circumferential strain maps (Ecc) derived from standard cine MRI videos.}
    \label{fig:cine-example}
\end{figure*}

\begin{figure*}[!ht]
    \centering
    \includegraphics[width=0.95\linewidth]{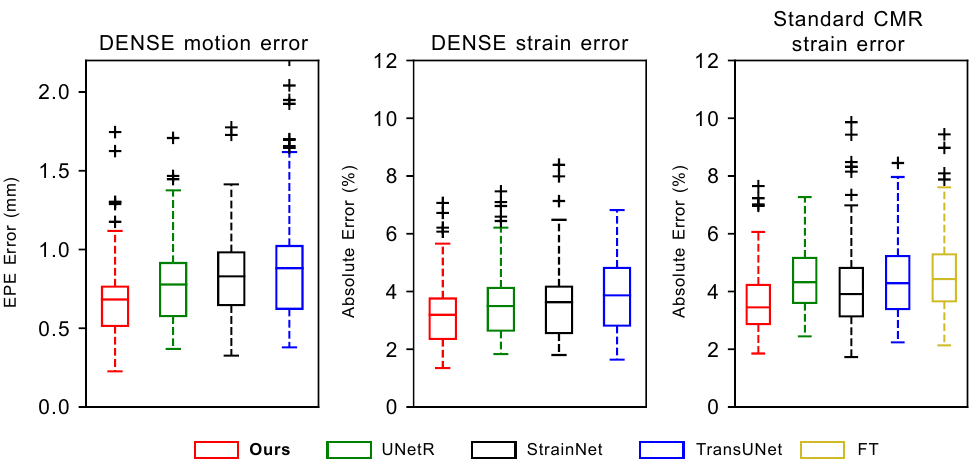}
    \caption{Left to right: a comparison of displacement field EPE error on DENSE; predicted segmental circumferential strain error on DENSE; and predicted segmental circumferential strain error on standard cine CMRs from our model vs. all baselines.}
    \label{fig:boxplots}
\end{figure*}

Similarly, the Fig.~\ref{fig:cine-example} (test on standard cine CMRs) visualizes the predicted motion fields and the circumferential (Ecc) strain  maps for all models compared to paired DENSE dataset. Note that the DENSE myocardium contours are slightly different from cine CMRs due to a different scanning time. 

Fig.~\ref{fig:boxplots} displays a quantitative comparison of displacement field EPE error and segmental circumferential strain error between our model and all baselines. The left two panels demonstrate the testing results on DENSE dataset, indicating that our method significantly outperforms the baselines in terms of both displacement and circumferential strain error.
The right panel shows the testing results on the cine CMRs dataset. It shows that our method consistently achieves superior performance of myocardial strain quality over all baselines.

\begin{figure}[!h]
    \centering
    \includegraphics[width=1.0\linewidth]{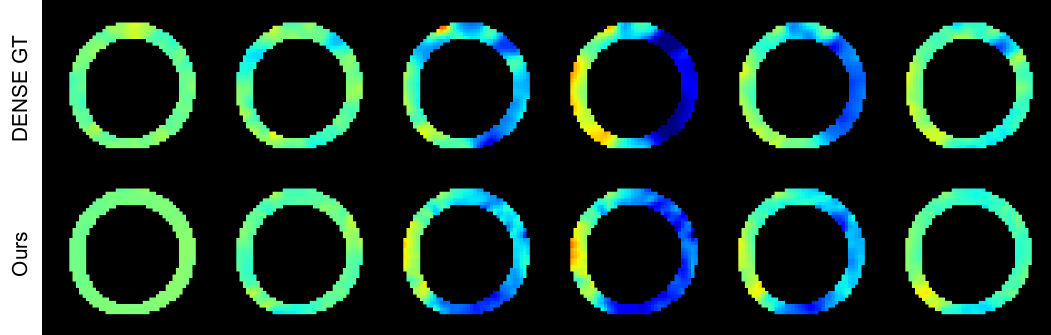}
    \caption{A comparison of time-sequential myocardial strain generated by DENSE vs. our model throughout the cardiac cycle.}
    \label{fig:over-frames}
\end{figure}
Fig.\ref{fig:over-frames} illustrates a comparison between circumferential strain computed from DENSE and our predictions across evenly sampled time frames throughout the cardiac cycle. The visual similarity between the top and bottom rows demonstrates that our method effectively captures the strain patterns at various phases of cardiac motion. 



\section{Conclusion}
In this paper, we introduced LaMoD, a novel Latent Motion Diffusion model that predicts highly accurate motions/strain from standard CMR videos. Our approach effectively addresses the challenges of detecting subtle myocardial movements, particularly in intramyocardial regions, by leveraging a pre-trained registration network and a probabilistic latent diffusion model in the latent motion space guided by DENSE CMRs. Experimental results demonstrate that LaMoD outperforms existing methods in motion prediction and strain generation accuracy. Our work has great potential to improve cardiac disease assessment based on strain, as well as treatment planning without requiring additional DENSE scans; hence ultimately improving patient care. Future work will focus on further validating the model's generalizability across diverse patient populations and clinical environments. \\

\noindent \textbf{Acknowledgement.} This work was supported by NSF CAREER Grant 2239977 and NIH 1R21EB032597. \\

\noindent {\bf Disclosure of Interests.} The authors have no competing interests to declare that are relevant to the content of this article.
\bibliographystyle{splncs04}
\bibliography{refs}
%




\end{document}